\documentclass[journal]{IEEEtran}

\usepackage{amsmath,amssymb,amsfonts}
\usepackage{booktabs}
\usepackage{multirow}
\usepackage{graphicx}
\usepackage{cite}
\usepackage{array}
\usepackage{tabularx}
\usepackage{xcolor}
\usepackage{url}
\usepackage[hidelinks]{hyperref}
\hypersetup{
  pdftitle={Cross-Fitted Residual Utility for Primary-Preserving Cognitive Decision Correction in Automatic Modulation Classification},
  pdfauthor={Linzhuo Han, Zongyong Cui, and Houbiao Li},
  pdfkeywords={automatic modulation classification; cognitive receiver; out-of-fold learning; decision correction; risk-aware routing}
}

\graphicspath{{figures_tccn/}}
\newcommand{\RMLA}{RML2016.10A}
\newcommand{\RMLB}{RML2016.10B}
\newcommand{\HISAR}{HisarMod2019.1}
\newcommand{\OOF}{\mathrm{OOF}}

\newcommand{\Prob}{\mathbf{p}}
\newcommand{\Ind}{\mathbb{I}}
\DeclareMathOperator*{\argmax}{arg\,max}

\title{Cross-Fitted Residual Utility for Primary-Preserving Cognitive Decision Correction in Automatic Modulation Classification}

\author{Linzhuo Han, Zongyong Cui, and Houbiao Li%
\thanks{Linzhuo Han and Houbiao Li are with the School of Mathematical Sciences, University of Electronic Science and Technology of China, Chengdu, China. Corresponding author: lhb0189@uestc.edu.cn or lihoubiao0189@163.com.}%
\thanks{Zongyong Cui is with the School of Information and Communication Engineering, University of Electronic Science and Technology of China, Chengdu, China.}%
\thanks{Source code and frozen evaluation artifacts for the reported experiments are available at \url{https://github.com/915213013-star/KAN-Fourier-RC}.}}

\begin{document}

\maketitle

\begin{abstract}
Automatic modulation classification research has largely emphasized representation accuracy, but a cognitive receiver must also decide when heterogeneous evidence justifies overriding a trusted default prediction. We study this post-inference problem through cross-fitted residual utility and a primary-preserving cognitive decision policy. A structured KAN-Fourier classifier supplies the default probability, while neural and non-neural candidates provide observable evidence. Candidate-specific residual utility is learned from train-split out-of-fold predictions, and a disjoint validation split freezes action thresholds, approved transitions, conditional routes, and a unified risk mask before held-out evaluation. On \RMLA, \RMLB, and \HISAR, the complete system improves overall accuracy from 63.632\% to 66.332\%, 65.161\% to 66.168\%, and 77.769\% to 79.867\%, respectively. Controlled comparisons show that the isolated utility target does not uniformly dominate alternative out-of-fold meta-learners; the consistent gain comes from the complete evidence-and-action policy. Paired bootstrap and Holm-corrected McNemar analyses support the controlled gains. A frozen-policy stress test under carrier-frequency offset, I/Q imbalance, and synthetic Rayleigh/Rician fading yields positive gains in all 11 conditions, with every paired 95\% confidence interval above zero.
\end{abstract}

\begin{IEEEkeywords}
automatic modulation classification, cognitive receiver, out-of-fold learning, decision correction, risk-aware routing
\end{IEEEkeywords}

\section{Introduction}
\label{sec:introduction}

\IEEEPARstart{A}{utomatic} modulation classification (AMC) is a central perception function in spectrum monitoring, adaptive demodulation, and cognitive radio receivers \cite{hameed2009,zhang2022survey}. Modern neural classifiers have substantially advanced representation-level accuracy through raw-I/Q learning, temporal attention, complex-valued processing, and communication-informed features \cite{oshea2016,rajendran2018,xu2020mcldnn}. Recent work has also examined multimodal fusion, compact models, semi-supervised learning, and adaptation to changing signal distributions \cite{li2024cvtrn,shao2025iqformer,cppcnet,guo2024semiamr,zhang2024osda,ke2025gignet}. These advances improve the classifier itself. They do not remove a second receiver-level question: once several heterogeneous predictors have produced probability vectors, which disagreement, if any, contains enough evidence to justify overriding a trusted default decision? This paper studies that post-inference question rather than claiming a universally strongest feature extractor.

Unconditional averaging or stacking can improve mean accuracy, but it may also replace a correct primary decision with an erroneous auxiliary prediction. This risk is pronounced when an auxiliary predictor is weaker on average yet complementary for a small subset of modulation classes or signal conditions. Dynamic classifier selection addresses related competence questions \cite{cruz2018}, while reject-option classification formalizes the cost of withholding a decision \cite{chow1970}. AMC deployment instead requires a retain-or-correct action: the receiver should preserve its primary prediction unless candidate-specific evidence indicates that a correction is likely to rescue an error without causing disproportionate harm.

We cast this decision as a leakage-controlled cognitive action problem. The receiver first produces a structured primary probability and a pool of heterogeneous candidate probabilities. Train-split cross-fitting then supplies out-of-fold (OOF) evidence for learning each candidate's expected residual utility relative to the primary model. A disjoint validation split selects and freezes action thresholds, optional probability mixing, approved class transitions, and risk masks. Held-out samples are processed only after the full configuration is fixed. Inference consequently follows the cognitive chain
\begin{equation}
\begin{aligned}
\text{perception}&\ \longrightarrow\ \text{OOF residual reasoning}\\
&\ \longrightarrow\ \text{risk-aware action selection}.
\end{aligned}
\label{eq:cognitive_chain}
\end{equation}
The action set explicitly includes \emph{retain primary}; correction is not the default.

The primary classifier, KAN-Fourier, combines signal-aligned structural priors, compact statistics, and temporal encoding. IQCC-Former and low-depth statistical, pairwise, and geometry-inspired predictors provide candidate evidence rather than global replacements. These modules are useful because their errors are different, not because every candidate is individually stronger. The proposed residual-correction layer is therefore model-agnostic in principle, although this paper evaluates one fixed predictor pool.

The main contributions are:
\begin{enumerate}
\item We formulate multi-predictor AMC as primary-preserving cognitive action selection, in which retaining the default decision is an explicit action and every correction is evaluated relative to the primary classifier.
\item We develop leakage-controlled OOF residual evidence and an independently validation-frozen policy that combines candidate-specific utility estimates, approved transitions, conditional routes, and a unified risk mask. Test labels are absent from every optimization objective.
\item We evaluate the complete system on two RadioML benchmarks and the long-sequence \HISAR{} dataset using strong OOF decision baselines, paired significance tests, action-level attribution, partition sensitivity, synthetic receiver/channel perturbations, and deployment-complexity accounting.
\end{enumerate}

The experiments also delimit the claim. The isolated rescue-minus-harm objective does not uniformly dominate OOF linear stacking, OOF XGBoost stacking, or OOF candidate competence. The repeatable advantage is obtained by the complete residual-evidence and validation-frozen action-policy design. Accordingly, the paper does not claim a universal meta-learning objective or an unconditional AMC accuracy state of the art.

\section{Cognitive Receiver Model and Design Requirements}
\label{sec:receiver}

\subsection{Signal and Decision Interface}
Consider a non-cooperative receiver observing a complex baseband record of length $T$,
\begin{equation}
\begin{aligned}
x[n]&=e^{j(2\pi\Delta f nT_s+\varphi)}
\sum_{\ell=0}^{L_h-1}h_{\ell}s[n-\ell]+w[n],\\
&\hspace{4em} n=0,\ldots,T-1.
\end{aligned}
\label{eq:signal_model}
\end{equation}
where $s[n]$ is the unknown modulated waveform, $h_{\ell}$ is the $\ell$th channel tap, $L_h$ is the effective channel memory, $\Delta f$ and $\varphi$ denote residual carrier-frequency and phase offsets, $T_s$ is the sampling interval, and $w[n]$ is additive noise. The AMC module receives the sampled I/Q sequence but not the transmitter configuration, modulation label, or true SNR. Its output configures downstream processing such as demodulator selection, monitoring, and spectrum interpretation. A wrong modulation decision may therefore propagate into an incompatible receiver chain even when the underlying feature extractor is otherwise reliable.

Most representation-level AMC systems optimize a single mapping $f:\mathbb{R}^{2\times T}\rightarrow\Delta^{C-1}$. Our setting starts one step later. A deployed receiver already has a primary mapping $f_0$ and several independently trained evidence sources $f_k$. The design question is not whether all sources should be averaged, but whether a particular disagreement contains enough observable evidence to justify changing the primary decision. This distinction is important in practical systems: replacing a mature primary model may be costly, whereas adding a post-inference decision layer can improve selected failure modes without retraining the full receiver stack.

\subsection{Receiver-Level Design Requirements}
The receiver-level problem imposes four requirements. First, \emph{primary preservation}: retaining the default prediction must be an explicit action rather than the accidental result of a small fusion weight. Second, \emph{observable inference}: the policy may use probabilities, confidence, margins, disagreement, and blind-quality estimates, but not true SNR or test labels. Third, \emph{leakage control}: every residual target must be formed from OOF predictions, and all action thresholds must be frozen on a disjoint validation split. Fourth, \emph{action auditability}: every changed decision must be attributable to one mutually exclusive action family so that rescue, harm, and net gain can be checked exactly. These requirements distinguish the proposed receiver from an unrestricted ensemble and motivate the mathematical action model developed next.

\section{Related Work}
\label{sec:related}

\subsection{Representation-Level AMC}
Early learning-based AMC systems used convolutional and recurrent architectures on raw I/Q sequences \cite{oshea2016,rajendran2018,xu2020mcldnn}. Later work introduced Transformer-style context modeling and frame-wise embeddings \cite{chen2023feat}. Complex-valued architectures preserve phase relationships that can be obscured by independent real-valued processing \cite{liang2022complex}. CV-TRN and IQFormer further illustrate the value of coordinated I/Q processing and temporal interaction \cite{li2024cvtrn,shao2025iqformer}. Our IQCC-Former candidate adopts the broad dual-stream idea but uses its own frame construction, complex-correlation attention, fused descriptor, and OOF decision role.

Knowledge-guided systems incorporate communication structure directly into feature learning. Representative examples include compact partial convolutions and adaptive wavelet models \cite{cppcnet,quan2024alwnn}, coordinated I/Q models \cite{li2024cvtrn,shao2025iqformer}, and interpretable multi-domain predictors \cite{yu2026gamc}. Robustness has been approached from several complementary directions: corrected pseudo-labeling under limited supervision \cite{guo2024semiamr}, open-set domain adaptation in dynamic environments \cite{zhang2024osda}, graph-based representation learning \cite{ke2025gignet}, and test-time adaptation under unknown channel distortions \cite{shao2026tta}. These methods adapt or strengthen the representation. Our contribution begins after the candidate predictors have produced probability evidence and asks whether a frozen receiver should retain or revise the default decision. Because the cited studies use different partitions, preprocessing, and complexity conventions, we use them to establish methodological context rather than to construct a cross-paper accuracy leaderboard.

\subsection{Structured Priors and Decision-Level Learning}
KANs and Fourier-enhanced KANs offer compact nonlinear function parameterizations \cite{kan,fourierkan}. Quaternion processing and symmetric positive-definite (SPD) geometry offer signal-aligned priors for coupled I/Q features and covariance structure \cite{quaternion,spd}. We use these elements as a compact structural route within the primary model, not as separate universal contributions.

At the decision level, stacking learns a second-stage predictor from base-model outputs \cite{stacking}; XGBoost is a strong tabular learner for heterogeneous probability features \cite{xgboost}. Reject-option classification and dynamic selection address decision risk or local competence from complementary viewpoints \cite{chow1970,cruz2018}. Our distinction is the joint use of OOF primary-relative residual targets, a retain-primary action, candidate-specific routes, and a policy frozen on independent validation data. This construction seeks reliable conditional corrections rather than a new unconstrained global classifier.

\section{Problem Formulation}
\label{sec:problem}

Let $\mathbf{x}\in\mathbb{R}^{2\times T}$ denote an I/Q record and $y\in\{1,\ldots,C\}$ its modulation label. The primary classifier outputs $\Prob_0(\mathbf{x})\in\Delta^{C-1}$ and predicts
\begin{equation}
\widehat y_0(\mathbf{x})=\argmax_c p_{0,c}(\mathbf{x}).
\end{equation}
Candidate $k\in\{1,\ldots,K\}$ outputs $\Prob_k(\mathbf{x})$ and $\widehat y_k(\mathbf{x})$. The receiver chooses an action
\begin{equation}
a(\mathbf{x})\in\{\mathrm{retain},\mathrm{correct}(1),\ldots,
\mathrm{correct}(K)\}.
\label{eq:action_space}
\end{equation}

For sample $i$ and candidate $k$, define primary-relative residual utility
\begin{equation}
u_{ik}=
\Ind\{\widehat y_{ik}=y_i,\widehat y_{i0}\neq y_i\}
-
\Ind\{\widehat y_{ik}\neq y_i,\widehat y_{i0}=y_i\}.
\label{eq:utility}
\end{equation}
Thus $u_{ik}=+1$ is a rescue, $u_{ik}=-1$ is harm, and $u_{ik}=0$ leaves correctness unchanged. This target is deliberately local: it describes the consequence of candidate $k$ relative to the primary prediction and is not asserted to be an optimal universal meta-learning objective.

The deployed action may be a hard candidate replacement or a validated blend. For a finite coefficient set $\mathcal{A}_\alpha\subset[0,1]$, define
\begin{equation}
\Prob_{k,\alpha}=(1-\alpha)\Prob_0+\alpha\Prob_k,
\qquad \widehat y_{k,\alpha}=\argmax_c p_{k,\alpha,c},
\label{eq:action_probability}
\end{equation}
and the corresponding action-specific utility
\begin{equation}
 u_{i,k,\alpha}=\Ind\{\widehat y_{i,k,\alpha}=y_i\}
 -\Ind\{\widehat y_{i0}=y_i\}.
\label{eq:action_utility}
\end{equation}
The current implementation first estimates candidate-level residual evidence and lets the validation policy select the executed action. Equations~\eqref{eq:action_probability}--\eqref{eq:action_utility} make the distinction explicit: candidate quality and action quality coincide for $\alpha=1$ but need not coincide for softer blends.

For a deployed policy, let $N_{\rm changed}$, $N_{\rm rescue}$, and $N_{\rm harm}$ be the numbers of changed, rescued, and harmed held-out samples. We report
\begin{equation}
U_{\rm cond}=
\frac{N_{\rm rescue}-N_{\rm harm}}{N_{\rm changed}},
\label{eq:cond_utility}
\end{equation}
when $N_{\rm changed}>0$. Changed, rescue, and harm rates are expressed as percentages of all samples; $U_{\rm cond}$ conditions only on changed samples.

\textbf{Proposition 1:} For any deterministic retain-or-correct policy evaluated on a fixed sample set of size $N$,
\begin{equation}
\mathrm{Acc}_{\rm final}-\mathrm{Acc}_{0}
=\frac{N_{\rm rescue}-N_{\rm harm}}{N}.
\label{eq:gain_identity}
\end{equation}
\emph{Proof:} Unchanged samples contribute equally to both accuracies. Among changed samples, rescues add one correct decision and harms remove one; all other changes have zero net effect. Summing these mutually exclusive cases yields \eqref{eq:gain_identity}. \hfill$\square$

Equation~\eqref{eq:gain_identity} is an accounting identity, not a generalization guarantee. The learning problem is to estimate positive conditional utility without optimistic in-sample evidence and to freeze a policy that preserves this balance on unseen data.

\textbf{Proposition 2 (conditional retain-or-correct rule):} Let $\mathcal{A}$ contain the retain action $a=0$ with utility $u_0=0$ and a finite set of candidate actions. If
\begin{equation}
 r_a(\boldsymbol{\phi})=\mathbb{E}[u_a\mid\boldsymbol{\phi}]
\end{equation}
is known, then an action maximizing conditional expected accuracy gain is
\begin{equation}
 a^\star(\boldsymbol{\phi})=\argmax_{a\in\mathcal{A}} r_a(\boldsymbol{\phi}).
\label{eq:bayes_action}
\end{equation}
In particular, retaining the primary decision is optimal whenever $\max_{a\neq 0}r_a(\boldsymbol{\phi})\leq 0$.

\emph{Proof:} Conditional on $\boldsymbol{\phi}$, the expected change in correctness produced by action $a$ is exactly $r_a(\boldsymbol{\phi})$. Selecting the maximum therefore maximizes conditional expected gain; since $r_0=0$, retain is optimal when all correction actions have non-positive utility. \hfill$\square$

This proposition motivates the explicit retain action and positive-utility thresholds. It does not claim that the learned estimator is exact. If competence models estimated both $\Pr(\widehat y_k=y\mid\boldsymbol{\phi})$ and $\Pr(\widehat y_0=y\mid\boldsymbol{\phi})$ perfectly, their difference would equal candidate ERU. In finite data, however, direct residual labels and separately estimated competence can behave differently, which is why both are included as controlled baselines rather than treated as theoretically incomparable methods.

\textbf{Proposition 3 (decision regret under utility estimation error):} Let $a^\star$ maximize the true conditional utility and let $\widehat a$ maximize an estimate $\widehat r_a$. If
\begin{equation}
\max_{a\in\mathcal{A}}\left|\widehat r_a(\boldsymbol{\phi})-r_a(\boldsymbol{\phi})\right|\leq\varepsilon,
\label{eq:uniform_error}
\end{equation}
then
\begin{equation}
0\leq r_{a^\star}(\boldsymbol{\phi})-r_{\widehat a}(\boldsymbol{\phi})\leq 2\varepsilon.
\label{eq:regret_bound}
\end{equation}
Moreover, if a correction is accepted only when $\widehat r_{\widehat a}(\boldsymbol{\phi})>\tau$ with $\tau>\varepsilon$, then $r_{\widehat a}(\boldsymbol{\phi})>\tau-\varepsilon>0$.

\emph{Proof:} Optimality of $\widehat a$ for $\widehat r$ gives $\widehat r_{\widehat a}\geq\widehat r_{a^\star}$. Adding and subtracting the two estimates and applying \eqref{eq:uniform_error} yields the $2\varepsilon$ bound. The threshold statement follows from $r_{\widehat a}\geq\widehat r_{\widehat a}-\varepsilon$. \hfill$\square$

The bound does not assert that a finite XGBoost estimator satisfies a known $\varepsilon$; it explains why positive validation margins and conservative thresholds are natural safeguards when utility estimates are imperfect.

\section{Primary-Preserving Residual Correction}
\label{sec:method}

\subsection{System Overview}
\begin{figure*}[t]
\centering
\includegraphics[width=0.99\textwidth]{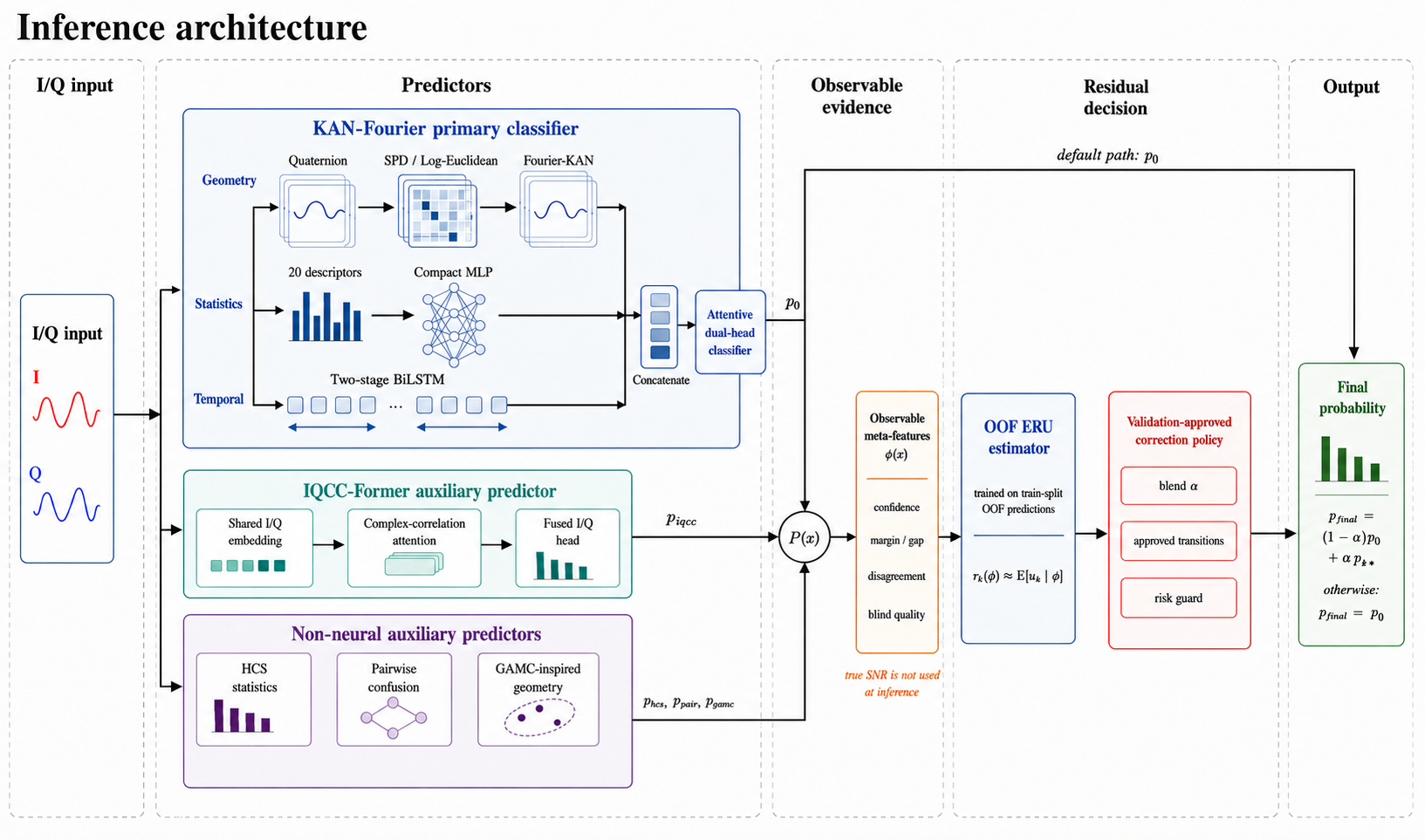}
\caption{Inference architecture of the primary-preserving cognitive receiver. The primary bypass implements the default retain action. Candidate probabilities are converted into observable evidence and OOF-ERU scores; only validation-approved actions can modify the final probability.}
\label{fig:inference}
\end{figure*}

\begin{figure*}[t]
\centering
\includegraphics[width=0.99\textwidth]{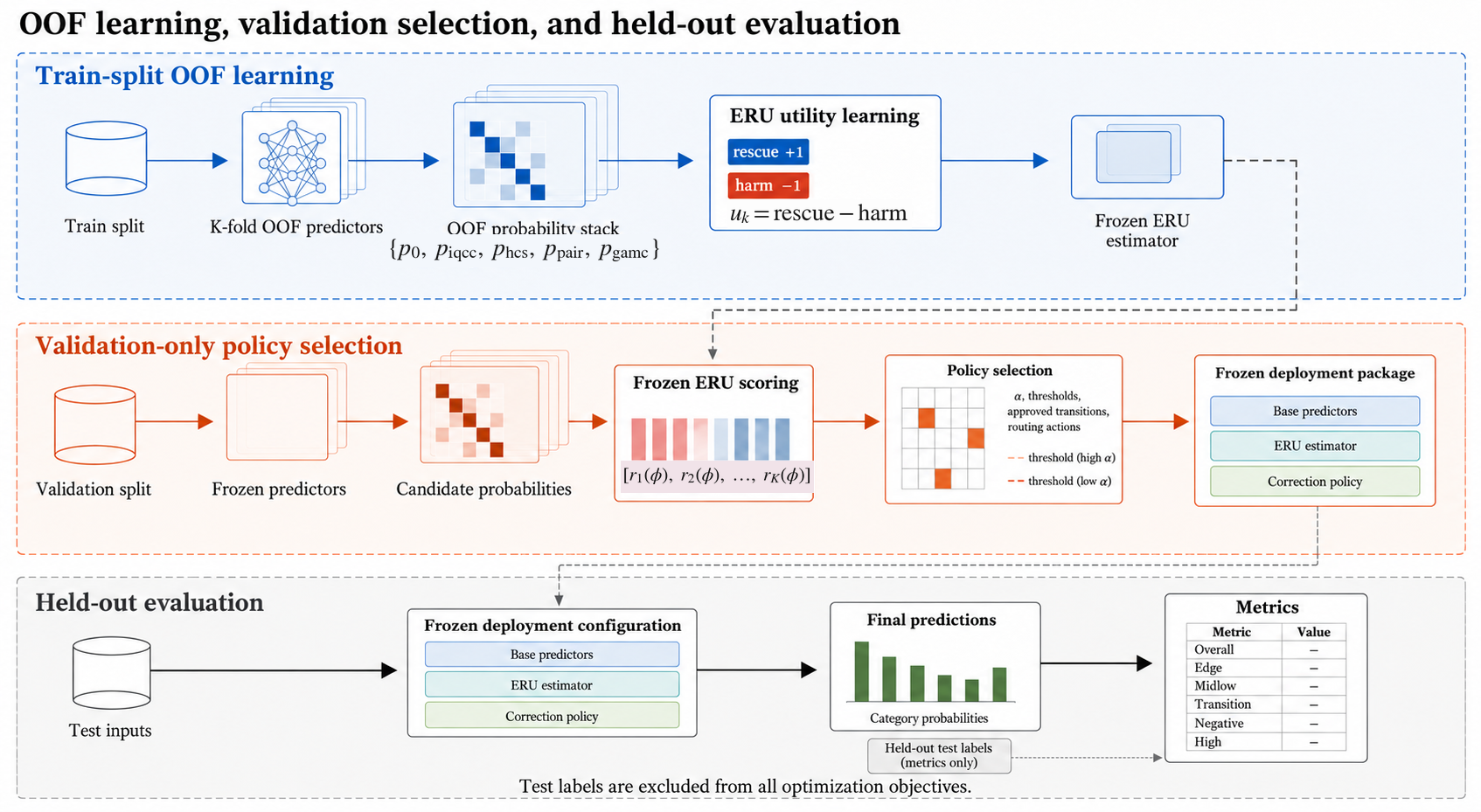}
\caption{Leakage-controlled OOF learning, validation-only policy selection, and held-out evaluation. Train-split OOF records fit the residual evidence models; the validation split freezes thresholds, action masks, and routing; test labels are used only to compute final metrics.}
\label{fig:protocol}
\end{figure*}

Figures~\ref{fig:inference} and~\ref{fig:protocol} separate two ideas that are often conflated in ensemble papers. Figure~\ref{fig:inference} is the deployed receiver: KAN-Fourier produces the default probability $\Prob_0$, heterogeneous predictors produce candidate probabilities, observable evidence is mapped to candidate-specific residual utility, and the validation-approved policy either retains $\Prob_0$ or applies one allowed correction. The upper bypass in Fig.~\ref{fig:inference} makes the default action explicit rather than implicit in a fusion coefficient.

Figure~\ref{fig:protocol} is the leakage-control protocol. OOF probabilities are generated only from models that did not train on the corresponding sample; the ERU estimators are fitted from those OOF records; policy parameters are selected only on validation data; and the resulting deployment package is evaluated on held-out inputs without optimization feedback. Separating these figures improves readability and clarifies that online cognition and offline protocol isolation are distinct contributions.

\subsection{Structured Primary Inference}
The primary classifier receives raw I/Q samples, a coupled structural representation, and 20 compact descriptors. Its three feature routes are
\begin{align}
\mathbf{z}_{\rm geo}&=h_{\rm KF}(\mathbf{x}),&
\mathbf{z}_{\rm stat}&=h_{\rm MLP}(\boldsymbol{\eta}),&
\mathbf{z}_{\rm time}&=h_{\rm BiLSTM}(\mathbf{x}),
\end{align}
where $h_{\rm KF}$ contains quaternion-based coupling, a regularized SPD covariance map, and Fourier-enhanced KAN layers; $\boldsymbol{\eta}$ contains moments and spectral descriptors. These operations are treated collectively as signal-aligned structural priors. The routes are concatenated, reweighted by channel attention, and passed to two internally routed logit heads:
\begin{align}
\mathbf{z}&=a_\theta(\mathbf{z}_{\rm geo}\Vert
\mathbf{z}_{\rm stat}\Vert\mathbf{z}_{\rm time}),\\
\Prob_0(\mathbf{x})&=\operatorname{softmax}\!
\left(\sum_{e=1}^{2}\pi_e(\mathbf{x})h_e(\mathbf{z})\right).
\label{eq:primary}
\end{align}
The two heads are internal to KAN-Fourier and should not be confused with the external correction candidates.

The RadioML primary contains 0.1973M trainable parameters and requires 2.848M FLOPs under the convention of two FLOPs per multiply--accumulate. For 1024-sample Hisar records, channel widths are adjusted to preserve long-range capacity without scaling cost linearly in every branch; the resulting primary contains 0.4676M parameters and requires 22.060M FLOPs.

\subsection{Auxiliary Evidence Pool}
IQCC-Former is the principal neural candidate. It retains separate I and Q frame streams, shares their embedding, and applies complex-correlation attention before a fused I/Q head. Its design is motivated by coordinated I/Q processing \cite{li2024cvtrn}, but its frame geometry, correlation operator, fused representation, training objective, and use as an OOF correction candidate are specific to this work. It outputs $\Prob_{\rm iqcc}(\mathbf{x})$ rather than replacing the primary model globally.

Three low-depth non-neural sources provide different evidence. The high-confidence statistical (HCS) source maps fixed moments, histograms, and spectral descriptors to class probabilities. The pairwise source specializes in recurrent primary--candidate confusion pairs. The GAMC-inspired source combines statistical, higher-order, Cartesian/polar, and graph-spectral summaries with shallow trees, adopting the multi-domain evidence principle of \cite{yu2026gamc} without using true SNR at inference. For \HISAR, a compact long-spectral-lite network adds a transform-domain probability source. Some candidates have lower standalone accuracy than KAN-Fourier; their role is justified only on subsets for which their conditional residual evidence is positive.

Table~\ref{tab:sources} summarizes the deployed evidence pool. The pool is deliberately heterogeneous. IQCC-Former uses 16-sample frames with stride 8, a 64-dimensional token space, and three four-head complex-correlation blocks in the RadioML configuration. Its I and Q streams share the frame embedding and attention projections before the fused head combines the two streams, their magnitude, and their element-wise product. This design makes its errors different from the geometry/statistics/temporal routes used by KAN-Fourier.

The non-neural sources are not decorative side information. In the formal RadioML implementation, the HCS map contains a high-confidence descriptor set, while the GAMC-inspired route combines statistical, spectral, higher-order, histogram, and fixed-size graph-spectral summaries. The graphs use a fixed node budget so that descriptor width does not grow with the raw sequence length. Pairwise models are trained only for pre-specified recurrent confusion families; they are not a bank of one-vs-one classifiers over all class pairs. For Hisar, long-spectral-lite supplies a compact transform-domain view because the 1024-sample records contain structure that is not fully represented by the short-frame IQCC configuration.

\begin{table}[t]
\centering
\caption{Deployed evidence sources and their intended roles.}
\label{tab:sources}
\small
\setlength{\tabcolsep}{3.5pt}
\begin{tabularx}{\columnwidth}{p{1.65cm}p{2.0cm}X}
\toprule
Source & Main evidence & Decision role \\
\midrule
KAN-Fourier & Geometry, compact statistics, temporal context & Default probability $\Prob_0$ \\
IQCC-Former & Cross-I/Q temporal correlation & Neural correction candidate \\
HCS & High-confidence fixed descriptors & Statistical candidate \\
Pairwise & Recurrent class-pair confusion evidence & Specialized correction candidate \\
GAMC-inspired & Statistical, spectral, and graph summaries & Multi-domain tree candidate \\
Long-spectral-lite & Long-sequence transform evidence & Hisar-only candidate \\
\bottomrule
\end{tabularx}
\end{table}

Every source is trained or cached independently before the action policy is selected. Consequently, a candidate can be weaker as a standalone classifier yet remain useful for a small, recognizable subset. This is not an exception to the framework but its motivating case: the decision layer evaluates residual value relative to the current primary rather than ranking models only by global accuracy.

\subsection{OOF Residual Evidence}
Directly fitting a correction model to in-sample base predictions would overestimate candidate competence. We partition the model-training set into $M$ folds. For sample $\mathbf{x}_i$ in fold $m$, every probability used for utility learning is generated by a predictor trained without fold $m$:
\begin{equation}
\Prob_{ik}^{\OOF}=f_k^{(-m)}(\mathbf{x}_i).
\label{eq:oof}
\end{equation}
Candidate-specific OOF labels are then computed from \eqref{eq:utility}. The ERU estimator targets
\begin{equation}
r_k(\boldsymbol{\phi})=\mathbb{E}\!\left[u_k\mid\boldsymbol{\phi}\right]
\label{eq:eru}
\end{equation}
and is fitted by the generic regularized OOF empirical-risk problem
\begin{equation}
\widehat r_k=\arg\min_{g\in\mathcal{G}}
\frac{1}{N_{\rm tr}}\sum_{i=1}^{N_{\rm tr}}
\ell\!\left(g(\boldsymbol{\phi}^{\OOF}_i),u_{ik}\right)
+\lambda\Omega(g),
\label{eq:eru_objective}
\end{equation}
where $\mathcal{G}$ is the shallow-tree function class and $\Omega$ is the XGBoost complexity penalty \cite{xgboost}. Because the target is candidate-specific, the policy can distinguish a useful pairwise correction from a harmful global switch even when both candidates predict the same class.

\subsection{Observable Meta-Features}
The vector $\boldsymbol{\phi}(\mathbf{x})$ contains only quantities available to a receiver at inference. For predictor $k$, let
\begin{align}
c_k&=\max_c p_{k,c},\qquad
m_k=p_{k,(1)}-p_{k,(2)},\label{eq:confidence_margin}\\
d_{0k}&=1-\frac{\Prob_0^{\mathsf T}\Prob_k}
{\|\Prob_0\|_2\|\Prob_k\|_2+\epsilon},
\label{eq:disagreement}
\end{align}
where $p_{k,(1)}$ and $p_{k,(2)}$ are the largest and second-largest class probabilities. The implemented meta-vector can be written compactly as
\begin{equation}
\boldsymbol{\phi}=\Big[\Prob_0^{\mathsf T},\ldots,\Prob_K^{\mathsf T},
\mathbf{c}^{\mathsf T},\mathbf{m}^{\mathsf T},
\mathbf{d}^{\mathsf T},\widehat{\mathbf{q}}_{\rm blind}^{\mathsf T}\Big]^{\mathsf T}.
\label{eq:meta}
\end{equation}
The compact blind-quality vector $\widehat{\mathbf{q}}_{\rm blind}$ is cross-fitted from observable signal statistics. True SNR is excluded from $\boldsymbol{\phi}$ and is used only for post-evaluation diagnostics.

For controlled baselines, OOF XGBoost stacking receives the same candidate pool and meta-features but directly predicts the final class. OOF candidate competence instead predicts candidate correctness and selects the most competent source. Isolated OOF ERU uses the score in \eqref{eq:eru} without the complete downstream action policy. This control isolates the effect of the objective from the effect of policy design.

\subsection{Validation-Frozen Cognitive Actions}
On the disjoint validation split, the system searches a finite action set comprising retention, direct candidate adoption, optional probability blending, approved class transitions, pairwise corrections, statistical/GAMC corrections, and a late conditional route. For a selected candidate $k^\star$,
\begin{equation}
\Prob_{\rm final}=
\begin{cases}
(1-\alpha)\Prob_0+\alpha\Prob_{k^\star},
& a=\mathrm{correct}(k^\star),\\
\Prob_0,& a=\mathrm{retain}.
\end{cases}
\label{eq:final_probability}
\end{equation}
The coefficient $\alpha$, utility thresholds, transition masks, action priority, and the unified risk mask are selected on validation data and then frozen. Because the three datasets use stage-specific executable selection rules, we write their common structure without imposing artificial shared constraints. For dataset $d\in\{\mathrm{10A},\mathrm{10B},\mathrm{Hisar}\}$, let $\Theta_d^{\rm adm}$ be the finite set of configurations that satisfy the pre-specified admissibility checks in the corresponding frozen pipeline, and let $S_{d,\rm val}$ denote its implemented validation score. Policy selection is summarized by
\begin{equation}
\widehat{\theta}_d
=\arg\max_{\theta\in\Theta_d^{\rm adm}}
S_{d,\rm val}(\theta).
\label{eq:policy_selection}
\end{equation}
Equation~\eqref{eq:policy_selection} is a faithful abstraction of the dataset-specific search rather than a claim that all three implementations share one scalar constraint set. The candidate set, scoring rule, and search range are fixed before held-out evaluation. The search includes blending and hard adoption; on \RMLA{} validation selects $\alpha=1$, so blending has no independent gain there. Transition actions may have zero activation when validation evidence is insufficient.

An action is \emph{retained} when no candidate satisfies its trigger. A \emph{blocked/reverted} action is a candidate proposal rejected by the frozen risk mask. The mask includes the former harm-guard logic; because the isolated guard changes no \RMLA{} prediction and is not separately identifiable on the other datasets, it is not presented as an independent contribution.

Base predictors are fitted on model-training folds, residual estimators are learned from train-split OOF records, and policy parameters are selected on the independent validation split. Held-out labels enter none of these objectives. The same separation is enforced for the proposed system and every controlled decision-level baseline.

\section{Experimental Protocol}
\label{sec:protocol}

\subsection{Datasets and Partitions}
\RMLA{} contains 220,000 records from 11 modulation classes over 20 SNR levels from $-20$ to 18~dB in 2-dB steps; each record contains 128 complex samples \cite{oshea2016}. A fixed stratified split allocates 176,000 samples to model training, 22,000 to policy validation, and 22,000 to held-out evaluation. \RMLB{} contains 1.2 million length-128 records from 10 classes on the same SNR grid; the corresponding 80/10/10 split contains 960,000, 120,000, and 120,000 samples.

\HISAR{} provides an official storage split with 520,000 training and 260,000 test records, 26 modulation classes, 20 SNR levels, and 1024 complex samples per record \cite{tekbiyik2020hisar}. We reserve 78,000 official-training records for policy validation and use the remaining 442,000 for model learning. Three folds of 147,680, 147,160, and 147,160 samples generate neural OOF probabilities. The official test storage is excluded from checkpoint, feature, threshold, and policy selection.

For RadioML, neural probability sources use three-fold OOF prediction caches, while compact quality and tree estimators use five-fold cross-fitting where specified. All records share aligned sample indices across sources. Neural models are optimized with AdamW and cosine learning-rate decay; checkpoint selection is performed on the appropriate held-out training fold. Batch sizes are 128 for RadioML and 256 for Hisar. The released package records the data indices, fold membership, checkpoint identifiers, model seeds, and frozen policy used for the reported runs; held-out labels are used only for final evaluation.

\subsection{Frozen-Policy Perturbation Test}
To probe receiver-level robustness without retraining or policy re-selection, we re-execute the frozen Hisar inference package under 11 synthetic conditions: a clean control, normalized carrier-frequency offsets (CFOs) of $\pm0.001$ and $\pm0.003$, mild and severe I/Q-imbalance perturbations of both signs, and synthetic Rayleigh and Rician fading. The exact gain/phase perturbation parameters, random seeds, and channel-generation settings are recorded in the released stress-test script. All base models, ERU estimators, thresholds, routing priorities, and masks remain fixed. Hence this experiment measures the behavior of the deployed decision mechanism rather than adaptation to the test condition.

The clean control is produced by the same re-inference pipeline as the perturbed conditions and yields 77.898\% for the Primary and 80.002\% for Frozen RC. These values are used only as an internal control for the stress test. The archived benchmark results, 77.769\% and 79.867\%, remain the formal Hisar results reported elsewhere in the paper. The small differences of 0.129 and 0.135~pp are consistent with FP32 execution, CUDA nondeterminism, batch partitioning, and re-execution of the inference chain; they are not used to replace or tune the archived result. Confidence intervals use 10,000 paired bootstrap replicates on the aligned stress-test predictions.

\subsection{Baselines and Metrics}
Decision baselines use the same candidate probabilities, OOF records, validation split, and held-out sample order:
\begin{itemize}
\item \emph{OOF linear stacking}: multinomial linear fusion.
\item \emph{OOF XGBoost stacking}: direct final-class prediction.
\item \emph{OOF candidate competence}: candidate-correctness estimation and selection.
\item \emph{Isolated OOF ERU}: candidate utility estimation without the complete action policy.
\item \emph{Full ERU-RC}: the complete validation-frozen policy.
\end{itemize}
We report overall accuracy, changed, rescue, harm, net gain, and conditional utility. Region metrics group low-SNR and transition behavior for diagnosis only. Displayed accuracies are rounded independently; all differences and net gains are computed from unrounded sample counts and reported in percentage points (pp).

\subsection{Controlled Fairness and Reproducibility}
All controlled decision-level methods receive the same stored probability records and are tuned under the same validation budget. OOF XGBoost stacking is not weakened by removing candidate probabilities or observable meta-features available to Full ERU-RC; its distinction is that it predicts the final class directly. Candidate competence uses the same evidence but optimizes candidate correctness rather than primary-relative utility. Isolated ERU uses the same residual target but omits the complete action policy. These controls separate three possible sources of improvement: the evidence pool, the meta-objective, and the downstream action design.

The paper reports fixed trained prediction records rather than claiming that full-system randomness has been exhausted. To maximize reproducibility, the released package records data indices, fold membership, model seeds, checkpoint identifiers, meta-model settings, and the frozen policy configuration. Paired tests quantify uncertainty over held-out samples; they do not substitute for a full retraining-seed study, and this boundary is stated explicitly in Section~\ref{sec:discussion}.

\subsection{Significance and Complexity}
Accuracy differences are evaluated with 10,000 paired bootstrap replicates. RadioML resampling is stratified by class and SNR; Hisar resampling uses class and SNR strata over the fixed official test storage. McNemar tests use the same paired predictions, and Holm correction is applied over the family of 15 pre-specified cross-dataset comparisons in the frozen analysis plan \cite{efron1979,mcnemar1947,holm1979}. Values below floating-point reporting precision are written as an inequality rather than $p=0$.

Neural complexity includes only probability sources retained at deployment; OOF fold models are training-only. One multiply--accumulate is counted as two FLOPs. Descriptor arithmetic, shallow-tree path comparisons, and constant-size policy lookups are reported separately because they are not equivalent to dense neural FLOPs.

\section{Results and Analysis}
\label{sec:results}

We first compare the complete policy with controlled decision-level baselines, then test paired significance, attribute the gains to concrete action families, and finally examine two complementary robustness questions: sensitivity to Hisar storage partitions and behavior under synthetic receiver/channel perturbations. The sequence moves from performance to evidence, mechanism, and stress testing.

\subsection{Controlled Decision-Level Comparison}
Table~\ref{tab:decision} compares six decision rules under a controlled protocol. Within each dataset, every learned rule receives the same aligned OOF probability pool, observable meta-features, validation split, and held-out sample order. This is stricter than comparing independently trained ensembles because the main difference is the decision objective and its downstream action policy.

\begin{table*}[t]
\centering
\caption{Controlled decision-level comparison. Conditional utility is $(N_{\rm rescue}-N_{\rm harm})/N_{\rm changed}$ in percent.}
\label{tab:decision}
\scriptsize
\setlength{\tabcolsep}{3.5pt}
\begin{tabular}{llrrrrrr}
\toprule
Dataset & Method & Overall & Changed & Rescue & Harm & Net gain (pp) & Conditional utility \\
\midrule
\multirow{6}{*}{\RMLA} & Primary & 63.632 & 0.000 & 0.000 & 0.000 & +0.000 & -- \\
& OOF Linear Stacking & 65.764 & 21.041 & 6.773 & 4.641 & +2.132 & 10.132 \\
& OOF XGBoost Stacking & 65.850 & 12.000 & 4.759 & 2.541 & +2.218 & 18.485 \\
& OOF Candidate Competence & 65.795 & 10.000 & 4.032 & 1.868 & +2.164 & 21.636 \\
& Isolated OOF ERU & 65.618 & 12.000 & 4.577 & 2.591 & +1.986 & 16.553 \\
& Full ERU-RC & \textbf{66.332} & 13.255 & 5.150 & 2.450 & \textbf{+2.700} & 20.370 \\
\midrule
\multirow{6}{*}{\RMLB} & Primary & 65.161 & 0.000 & 0.000 & 0.000 & +0.000 & -- \\
& OOF Linear Stacking & \textbf{66.171} & 8.555 & 2.650 & 1.640 & +1.010 & 11.806 \\
& OOF XGBoost Stacking & 65.860 & 37.797 & 9.420 & 8.721 & +0.699 & 1.850 \\
& OOF Candidate Competence & 65.928 & 1.998 & 1.080 & 0.312 & +0.767 & 38.423 \\
& Isolated OOF ERU & 65.939 & 1.944 & 1.052 & 0.273 & +0.778 & 40.034 \\
& Full ERU-RC & 66.168 & 8.000 & 2.983 & 1.976 & +1.008 & 12.594 \\
\midrule
\multirow{6}{*}{\HISAR} & Primary & 77.769 & 0.000 & 0.000 & 0.000 & +0.000 & -- \\
& OOF Linear Stacking & 78.651 & 12.592 & 3.781 & 2.899 & +0.882 & 7.001 \\
& OOF XGBoost Stacking & 79.744 & 12.167 & 4.605 & 2.630 & +1.975 & 16.229 \\
& OOF Candidate Competence & 79.016 & 4.061 & 2.135 & 0.888 & +1.247 & 30.707 \\
& Isolated OOF ERU & 79.257 & 6.715 & 2.982 & 1.494 & +1.488 & 22.153 \\
& Full ERU-RC & \textbf{79.867} & 10.433 & 4.204 & 2.107 & \textbf{+2.098} & 20.106 \\
\bottomrule
\end{tabular}

\vspace{1pt}
\parbox{0.98\textwidth}{\scriptsize Displayed accuracies are rounded independently; net gains use exact sample counts. On \RMLB, OOF Linear Stacking and Full ERU-RC differ by only $-0.0025$~pp (three samples); their paired 95\% CI is $[-0.1167,0.1133]$~pp and Holm-adjusted exact McNemar $p=0.9782$.}
\end{table*}

KAN-Fourier-RC raises overall accuracy from 63.632\% to 66.332\% on \RMLA, from 65.161\% to 66.168\% on \RMLB, and from 77.769\% to 79.867\% on \HISAR. The corresponding count-derived gains are 2.700, 1.008, and 2.098~pp. The complete system exceeds OOF XGBoost stacking by 0.482, 0.308, and 0.123~pp and exceeds OOF candidate competence by 0.536, 0.240, and 0.851~pp. On \RMLB, OOF linear stacking and Full ERU-RC reach 66.171\% and 66.168\%, respectively. Their paired difference is $-0.003$~pp (95\% CI $[-0.117,0.113]$), and the exact McNemar test detects no significant difference after Holm correction ($p_{\rm adj}=0.978$). They are therefore statistically indistinguishable under the reported protocol.

The isolated residual-utility objective is not uniformly strongest: it trails both OOF XGBoost stacking and OOF candidate competence on \RMLA, is competitive on \RMLB, and exceeds linear stacking but not XGBoost stacking on \HISAR. The result supports a narrower conclusion. Primary-relative utility is useful for constructing candidate actions, but consistent held-out improvement comes from combining residual evidence with candidate-specific routes and a validation-frozen policy.

\subsection{Controlled Gain Visualization}
\begin{figure*}[t]
\centering
\includegraphics[width=0.96\textwidth]{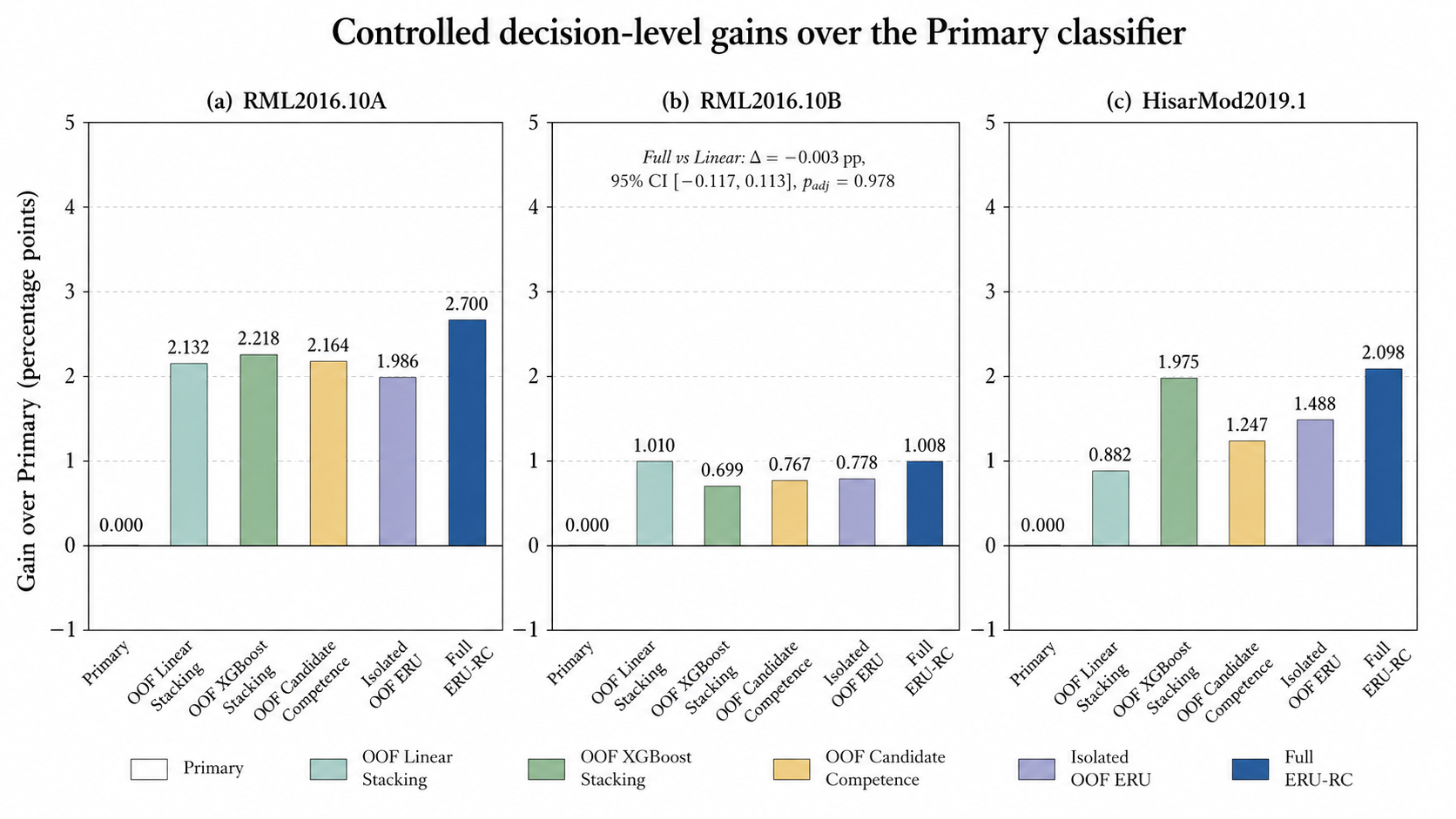}
\caption{Controlled decision-level gains relative to the Primary classifier. Every bar starts at zero, so its height is proportional to the reported percentage-point gain. Exact overall accuracies, changed rates, rescue, harm, and conditional utility are listed in Table~\ref{tab:decision}.}
\label{fig:comparison}
\end{figure*}

Figure~\ref{fig:comparison} plots gain over the Primary classifier rather than drawing absolute-accuracy bars from a truncated nonzero baseline. Bars show absolute percentage-point gains from a zero baseline, which makes the decision-layer effect directly comparable across datasets. Full ERU-RC obtains the largest gain on \RMLA{} and \HISAR{} and is statistically indistinguishable from linear stacking on \RMLB{}. The varying order of the intermediate methods reinforces the central result: no isolated OOF objective dominates across all datasets, whereas the complete policy provides the strongest or statistically tied operating point.

\subsection{Regional Behavior and Primary Preservation}
Overall accuracy alone does not reveal whether a decision layer improves difficult signal conditions at the expense of reliable ones. Table~\ref{tab:regional} reports the same frozen predictions over pre-specified SNR regions. On \RMLA, the full policy gains 3.000~pp at the two edge SNRs, 5.091~pp in the mid-low region, 3.727~pp over the transition region, and 3.937~pp across all negative SNRs. The high-SNR gain remains positive at 1.464~pp. \RMLB{} and \HISAR{} exhibit the same broad direction.

\begin{table*}[t]
\centering
\caption{Regional held-out accuracy of the Primary classifier and Full ERU-RC. SNR is used only for post-evaluation reporting.}
\label{tab:regional}
\small
\setlength{\tabcolsep}{5pt}
\begin{tabular}{llrrrrr}
\toprule
Dataset & Method & Edge & Mid-low & Transition & Negative & High \\
\midrule
\multirow{2}{*}{\RMLA} & Primary & 9.500 & 14.364 & 58.255 & 34.918 & 92.345 \\
& Full ERU-RC & \textbf{12.500} & \textbf{19.455} & \textbf{61.982} & \textbf{38.855} & \textbf{93.809} \\
\midrule
\multirow{2}{*}{\RMLB} & Primary & 10.650 & 18.450 & 60.477 & 37.270 & 93.052 \\
& Full ERU-RC & \textbf{12.675} & \textbf{21.492} & \textbf{61.090} & \textbf{38.827} & \textbf{93.510} \\
\midrule
\multirow{2}{*}{\HISAR} & Primary & 58.105 & 60.050 & 66.718 & 62.801 & 92.738 \\
& Full ERU-RC & \textbf{59.872} & \textbf{61.504} & \textbf{68.369} & \textbf{64.447} & \textbf{95.287} \\
\bottomrule
\end{tabular}
\end{table*}

These values are diagnostic rather than routing inputs: true SNR is never present in $\boldsymbol{\phi}$. The regional table explains why a low overall correction rate can still be useful--the policy concentrates actions where the primary error surface is structured and candidate disagreement is informative.

\subsection{Paired Statistical Evidence}
Table~\ref{tab:significance} reports paired uncertainty for the strongest pre-specified decision baselines. The six comparisons with OOF XGBoost stacking or candidate competence have bootstrap intervals strictly above zero and remain significant after Holm correction. The additional \RMLB{} comparison with OOF linear stacking crosses zero and yields $p_{\rm adj}=0.978$, confirming statistical indistinguishability rather than merely numerical closeness. We do not report $p=0$; very small values are represented in scientific notation. The smaller Hisar margin over XGBoost stacking, 0.123~pp, remains supported because it is computed from 260,000 paired held-out decisions rather than from independently sampled aggregate accuracies.

\begin{table*}[t]
\centering
\caption{Selected paired significance results for Full ERU-RC. Confidence intervals use 10,000 paired bootstrap replicates stratified by class and SNR. The Holm adjustment is computed over 15 pre-specified cross-dataset comparisons; only the comparisons most relevant to the decision-level claim are displayed. Here $n_{10}$ counts samples correct only under Full ERU-RC and $n_{01}$ counts samples correct only under the comparator.}
\label{tab:significance}
\scriptsize
\setlength{\tabcolsep}{3.5pt}
\begin{tabular}{llccrrr}
\toprule
Dataset & Full ERU-RC versus & Difference (pp) & Stratified 95\% CI & $n_{10}$ & $n_{01}$ & Holm-adjusted $p$ \\
\midrule
\RMLA & OOF XGBoost Stacking & +0.482 & [0.236, 0.727] & 464 & 358 & $1.23\times10^{-3}$ \\
\RMLA & OOF Candidate Competence & +0.536 & [0.259, 0.818] & 582 & 464 & $1.23\times10^{-3}$ \\
\midrule
\RMLB & OOF Linear Stacking & $-0.003$ & [$-0.117$, 0.113] & 2680 & 2683 & 0.9782 \\
\RMLB & OOF XGBoost Stacking & +0.308 & [0.122, 0.499] & 8332 & 7962 & $7.68\times10^{-3}$ \\
\RMLB & OOF Candidate Competence & +0.240 & [0.123, 0.356] & 2711 & 2423 & $4.31\times10^{-4}$ \\
\midrule
\HISAR & OOF XGBoost Stacking & +0.123 & [0.071, 0.173] & 2501 & 2181 & $2.48\times10^{-5}$ \\
\HISAR & OOF Candidate Competence & +0.851 & [0.770, 0.930] & 7041 & 4829 & $5.67\times10^{-91}$ \\
\bottomrule
\end{tabular}
\end{table*}

The paired statistical conclusions apply only to methods reconstructed from the same frozen candidate evidence and evaluated on the same held-out samples. Results from protocol-different representation models remain contextual references and are not used as evidence for the decision-level claim.

\subsection{Evidence for the Primary-Model Role}
A residual correction system requires a designated default model, but the predictor with the highest standalone accuracy is not automatically the best primary. On \RMLA, IQCC-Former reaches 64.555\%, 0.923~pp above KAN-Fourier. In a reciprocal role experiment, the IQCC-Former-primary system reaches 65.800\%, whereas the KAN-Fourier-primary system reaches 66.332\%. Relative to their standalone models, the systems gain 1.245 and 2.700~pp, respectively. Primary selection therefore depends on the residual structure exposed to the candidate pool, not only on standalone accuracy.

A fixed-run implementation-sensitivity check under approximately matched parameter and FLOP budgets gives 63.523\% for a real multiscale-convolution replacement, 63.668\% for first-order pooling, and 53.900\% for a matched SiLU MLP, compared with 63.632\% for KAN-Fourier. The first two differences are too small to support independent component-level claims; quaternion coupling and SPD geometry are therefore presented as structural priors. The 9.732-pp drop of the SiLU replacement supports the periodic nonlinear route in this implementation, without claiming universal superiority over all alternative backbones. The reciprocal role experiment and this sensitivity check serve different purposes: the former justifies the operational default, whereas the latter identifies which design choices are strongly or weakly supported by the current fixed-run evidence.

\subsection{Where the Corrections Come From}
Table~\ref{tab:actions} audits mutually exclusive final actions. For each dataset, changed-action activation rates add to the reported changed rate. Individual gain entries are rounded independently, whereas each total is recomputed from exact sample counts; the displayed row sum may therefore differ from the displayed total by 0.001--0.002~pp. The largest \RMLA{} contribution is statistical/GAMC correction (1.127~pp), followed by basic residual replacement (0.764~pp) and the late conditional route (0.473~pp). \RMLB{} obtains smaller but distributed gains, whereas the Hisar improvement is driven by the long-sequence late route (0.893~pp), basic replacement (0.505~pp), and statistical/GAMC evidence (0.498~pp).

\begin{table*}[t]
\centering
\caption{Mutually exclusive action-family audit. Reported values are net-gain contributions in percentage points relative to the Primary classifier. Zero activation corresponds to zero contribution. Retain-primary actions have zero gain by definition and are omitted for compactness.}
\label{tab:actions}
\small
\setlength{\tabcolsep}{5pt}
\begin{tabular}{lccc}
\toprule
Action family & \RMLA{} gain (pp) & \RMLB{} gain (pp) & \HISAR{} gain (pp) \\
\midrule
Basic ERU candidate replacement & 0.764 & 0.450 & 0.505 \\
Validation-approved transition correction & 0.100 & 0.000 & 0.000 \\
Pairwise confusion correction & 0.236 & 0.045 & 0.200 \\
Statistical/GAMC candidate correction & 1.127 & 0.425 & 0.498 \\
Late-stage conditional route & 0.473 & 0.087 & 0.893 \\
\midrule
Total Full ERU-RC gain & \textbf{2.700} & \textbf{1.008} & \textbf{2.098} \\
\bottomrule
\end{tabular}

\vspace{1pt}
\parbox{0.96\textwidth}{\scriptsize Individual action contributions are rounded independently to three decimals, whereas each total is recomputed from exact sample counts. Consequently, the displayed row sum differs from the displayed total by 0.001~pp on \RMLB{} and 0.002~pp on \HISAR{}.}
\end{table*}

Panel (b) uses a different reference point and must not be read as final action attribution. Relative to isolated OOF ERU, the complete policy adds 0.714, 0.229, and 0.610~pp on \RMLA, \RMLB, and \HISAR. The Hisar increment is dominated by late routing and blocked/reverted proposals; the latter denotes proposals suppressed by the frozen risk policy, not a separately trained classifier. The \RMLB{} blocked/reverted term is slightly negative, illustrating that the final gain is attributable to coordinated action design rather than a monotonic accumulation of individually beneficial switches.

\subsection{Policy Progression and Coverage--Utility Balance}
The policy audit includes negative and null stages rather than presenting the complete route as a sequence of uniformly beneficial modules. On \RMLA, validation selects $\alpha=1$, so the optional blending action collapses to hard candidate adoption. Adding approved transitions changes overall accuracy by only $-0.032$~pp because the reduction in harm is offset by lost rescues, and the separately measured harm guard changes no prediction; its logic is therefore absorbed into the unified risk mask. These outcomes support the use of a finite validation-approved action set, but they are not claimed as independent accuracy improvements.

Proposition~1 also gives a useful operating-point decomposition,
\begin{equation}
 \Delta\mathrm{Acc}=\Pr(a\neq\mathrm{retain})\,
 \mathbb{E}[u_a\mid a\neq\mathrm{retain}],
\label{eq:coverage_utility}
\end{equation}
which is the product of correction coverage and conditional utility. The values in Table~\ref{tab:decision} show why neither factor should be optimized alone. On \RMLA, candidate competence has the highest conditional utility but covers fewer correctable errors, whereas Full ERU-RC accepts a broader set and obtains the largest net gain. On \RMLB, XGBoost stacking changes 37.798\% of all decisions but has only 1.850\% conditional utility; Full ERU-RC changes 8.000\% and obtains a larger gain. The complete policy is therefore best understood as a coverage--utility operating point rather than a universally superior scalar score.

\subsection{Policy Sensitivity to Hisar Storage Partitions}
\begin{figure*}[t]
\centering
\includegraphics[width=0.96\textwidth]{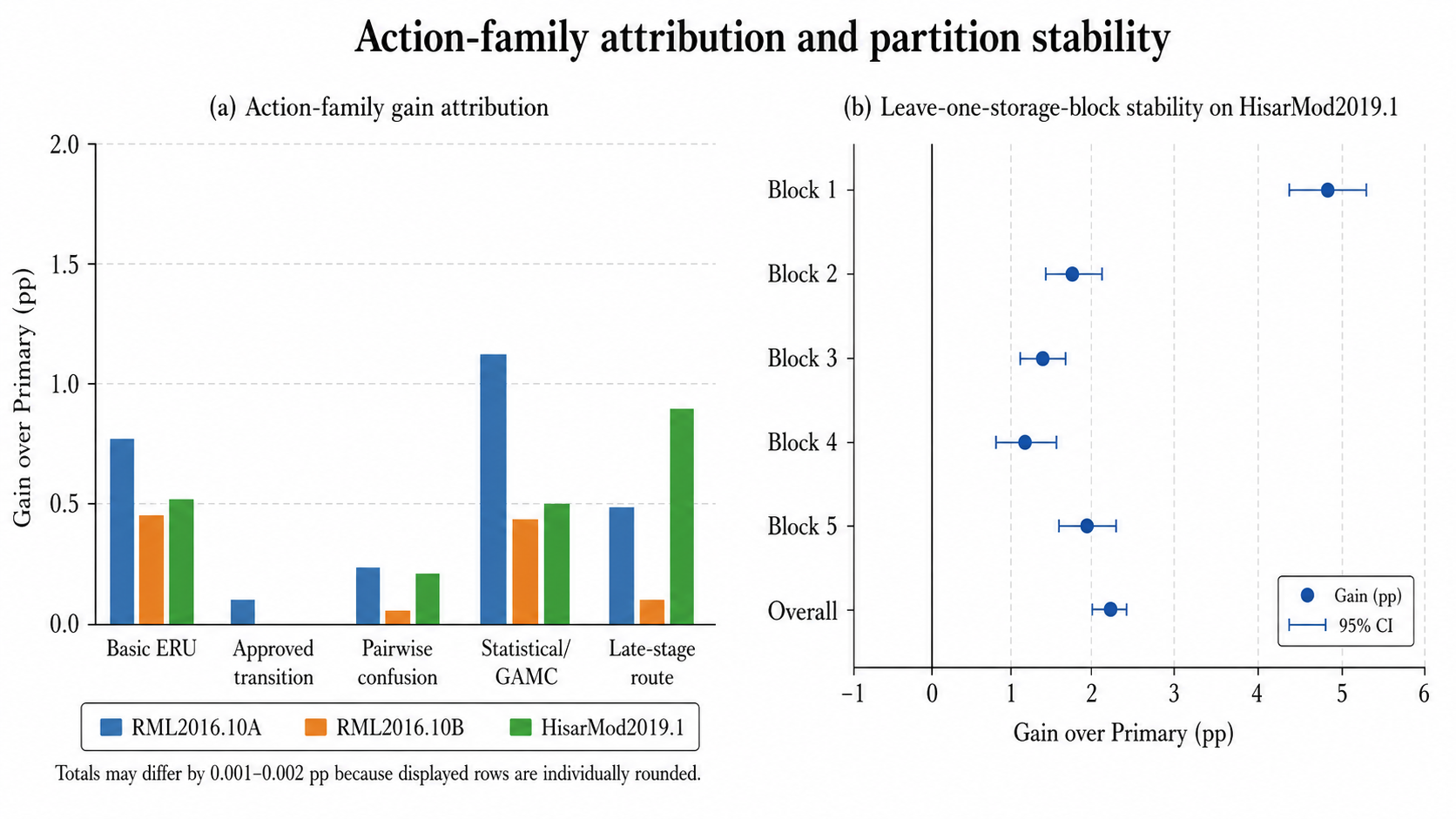}
\caption{Action audit and partition sensitivity. (a) Mutually exclusive action-family gains relative to Primary. (b) Hisar leave-one-storage-block sensitivity with paired 95\% bootstrap intervals. Storage blocks are storage partitions, not inferred physical channel labels.}
\label{fig:audit}
\end{figure*}

The processed Hisar files provide no authoritative per-sample physical-channel label. We therefore do not rename storage order as Rayleigh, Rician, Nakagami, or cross-channel transfer. Instead, Fig.~\ref{fig:audit}(b) reports a post-hoc leave-one-storage-block sensitivity analysis over five equal-sized test-storage partitions. Removing each block from the policy-selection diagnostic recovers the same frozen policy. Its gains on the five omitted blocks are 4.681, 1.635, 1.292, 1.115, and 1.765~pp, with paired 95\% intervals $[4.417,4.942]$, $[1.421,1.844]$, $[1.110,1.467]$, $[0.935,1.300]$, and $[1.556,1.975]$~pp.

Across the complete Hisar test storage, the gain is 2.098~pp with a 95\% interval of approximately $[2.003,2.190]$~pp; changed, rescue, and harm rates are 10.433\%, 4.204\%, and 2.107\%. Thus the policy-selection result is stable to deleting one storage partition, although the gain magnitude is heterogeneous. Because this analysis was conducted after the formal configuration had been frozen, it is a sensitivity analysis and did not participate in model or policy selection.

\subsection{Frozen-Policy Robustness to Receiver and Channel Perturbations}
\begin{figure*}[t]
\centering
\includegraphics[width=0.96\textwidth]{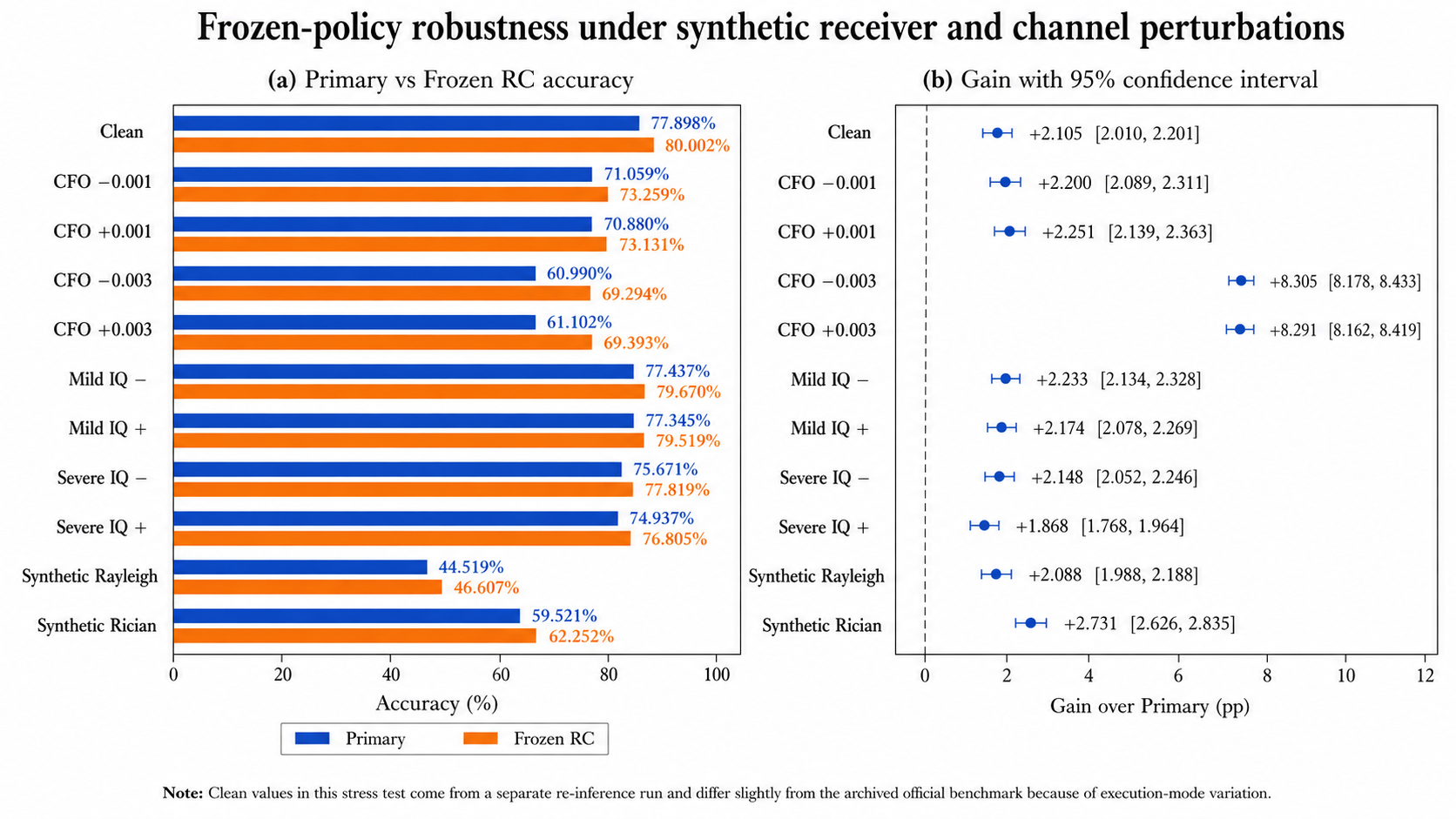}
\caption{Frozen-policy stress test on HisarMod2019.1. (a) Primary and Frozen RC accuracy under synthetic CFO, I/Q-imbalance, Rayleigh, and Rician perturbations. (b) Paired gain with 95\% bootstrap confidence intervals. The clean values are from the same re-inference run used for the stress test and do not replace the archived benchmark results.}
\label{fig:robustness}
\end{figure*}

Figure~\ref{fig:robustness} reports the frozen Hisar policy under the 11 stress-test conditions defined in Section~\ref{sec:protocol}. The paired gain is positive in every condition, and every 95\% confidence interval has a lower endpoint above zero. Under the clean re-inference control, Frozen RC improves 77.898\% to 80.002\% ($+2.105$~pp). Mild and severe I/Q imbalance retain gains between 1.868 and 2.233~pp, while the smaller CFOs retain gains of 2.200 and 2.251~pp. The larger CFOs reduce the Primary accuracy to approximately 61\%, but the frozen correction policy recovers more than 8.29~pp. This larger gain should be interpreted as increased residual headroom when the Primary deteriorates, not as immunity to CFO.

Synthetic Rayleigh and Rician fading also produce positive gains of 2.088 and 2.731~pp, respectively. Their absolute accuracies, 46.607\% and 62.252\% after correction, remain well below the clean case. The experiment therefore supports a narrower and useful claim: the frozen retain-or-correct mechanism continues to add value under the tested perturbations. It does not establish over-the-air invariance or transfer to arbitrary physical channels.

\subsection{Deployment Complexity}
Table~\ref{tab:complexity} reports the deployed neural-model complexity subtotal and separates it from training-only OOF copies. For length-128 RadioML inputs, the neural subtotal is 0.3694M parameters and 10.861M FLOPs. For length-1024 Hisar inputs, KAN-Fourier, IQCC-Former, and long-spectral-lite total 0.7262M parameters and 65.022M FLOPs.

\begin{table}[t]
\centering
\caption{Deployed neural-model complexity subtotal. Only neural probability sources retained at inference are included. CPU-side descriptors, shallow trees, and policy logic are disclosed separately and are not converted to dense-neural FLOPs.}
\label{tab:complexity}
\scriptsize
\setlength{\tabcolsep}{3.5pt}
\begin{tabular}{lcc}
\toprule
Configuration & Parameters (M) & FLOPs (M) \\
\midrule
RadioML primary (KAN-Fourier) & 0.1973 & 2.848 \\
RadioML deployed neural subtotal & 0.3694 & 10.861 \\
Hisar primary (KAN-Fourier) & 0.4676 & 22.060 \\
Hisar deployed neural subtotal & 0.7262 & 65.022 \\
\bottomrule
\end{tabular}
\end{table}

The non-neural side computes a fixed descriptor vector, graph summaries, shallow-tree paths, and a constant-size policy lookup. These operations are inexpensive relative to neural inference but depend on the tree backend and hardware; converting them to a dense-neural FLOP number would be misleading. The Hisar configuration is consequently compact relative to its three neural probability sources, but 65.022M FLOPs for a 1024-sample record should not be described as lightweight in an absolute embedded-receiver sense.

For RadioML, descriptor extraction uses fixed-size statistics and four 32-node graph spectra; per-sample arithmetic is approximately linear or FFT-linear in sequence length plus small fixed eigendecompositions. Tree inference consists mainly of threshold comparisons and memory-dependent branch traversal rather than dense multiply--accumulates. Neural FLOPs, descriptor structure, tree paths, and rule-table size are therefore disclosed separately. The complete heterogeneous pipeline cannot be summarized by one exact hardware-independent FLOP value.

Deployment cost must also be separated from offline cost. OOF construction trains fold copies and stores probability caches, but no fold copy is retained at inference. A future implementation could reduce average cost with a pre-gate that skips selected auxiliary neural sources for decisive primary predictions; the current system evaluates decision selectivity, not conditional neural execution.

\section{Discussion and Limitations}
\label{sec:discussion}

The experiments support a decision-level mechanism rather than an unconditional representation-level claim. A heterogeneous predictor can be weaker on average and still be useful when its residual value is positive on a recognizable subset. The new stress test strengthens this interpretation: without refitting any model or threshold, the frozen policy retains positive paired utility under all tested CFO, I/Q-imbalance, and synthetic fading conditions. OOF prediction is important because it removes the optimistic evidence that would arise if the action learner saw base-model training outputs. Independent validation then serves a different purpose: it freezes which proposed actions are acceptable. The explicit retain action makes the resulting behavior primary-preserving rather than merely another global ensemble.

The evidence also clarifies what is \emph{not} established. The rescue-minus-harm label alone is not universally superior to linear stacking, XGBoost stacking, or candidate-competence learning. Blending is part of the search space, but \RMLA{} selects hard candidate adoption, so it has no separate gain there. Validation-approved transitions can have zero activation, and the former harm guard is folded into one risk mask because its independent effect is not identifiable across all datasets. These outcomes reduce the number of claimed modules, but they strengthen the interpretation of the complete policy.

\subsection{Relation to Representation-Level AMC}
Recent compact, multimodal, semi-supervised, and adaptation-based AMC models address how to learn a stronger representation under their respective protocols. Our controlled evidence addresses a different question: how to use a fixed set of heterogeneous probabilities after inference without unnecessarily overwriting a reliable default. Cross-paper accuracies are therefore discussed only as methodological context; the decision-level claim is supported by baselines that share the same probability pool, validation data, and held-out sample order.

A stronger future primary classifier can be inserted into the same interface, but the OOF records, ERU estimators, and validation policy must then be rebuilt because residual utility is primary-relative. The method is thus a disciplined post-classification correction layer, not a substitute for continued progress in representation learning.

Several limitations remain. First, the paired bootstrap and McNemar analyses condition on frozen prediction records and do not quantify the variance of retraining every neural candidate. A controlled primary-seed study with a fixed auxiliary pool would provide targeted evidence about sensitivity to the primary error surface. Second, the candidate pool is engineered around the evaluated datasets; new signal families may require new evidence sources even though the correction interface is generic. Third, the Hisar neural subtotal is materially larger than the RadioML subtotal, and its 65.022M-FLOP long-sequence configuration is not yet a fully lightweight implementation. Fourth, shallow-tree latency and descriptor cost are implementation-dependent and have not been benchmarked on a common embedded platform. Fifth, external AMC studies use different splits and complexity conventions, and our results should not be read as a cross-protocol state-of-the-art claim. Finally, the storage-block analysis and synthetic perturbation test provide complementary stability evidence, but neither substitutes for over-the-air validation across independent transmitters, receivers, and propagation environments.

Future work will evaluate the frozen policy on streaming software-defined-radio captures, measure end-to-end latency and energy on shared hardware, and extend the current synthetic stress test to datasets with verified device, channel, and impairment labels. Long-sequence source compression, online drift detection, and uncertainty sets for candidate actions are also promising directions. Such extensions should preserve the central protocol requirement: model fitting, utility learning, and policy selection must remain separated from final evaluation.

\section{Conclusion}
\label{sec:conclusion}

This paper presented cross-fitted residual utility for primary-preserving cognitive decision correction in AMC. The receiver treats heterogeneous predictions as observable evidence, learns candidate-specific residual utility from train-split OOF predictions, and freezes a retain-or-correct policy on independent validation data. The complete KAN-Fourier-RC system improves its primary classifier by 2.700, 1.008, and 2.098~pp on \RMLA, \RMLB, and \HISAR, respectively. Controlled baselines show that the isolated utility target is not uniformly dominant; the stable benefit comes from the complete residual-evidence and risk-aware action design. Paired tests, action accounting, storage-partition sensitivity, and frozen-policy perturbation experiments make the source and limits of these gains explicit. Across all 11 synthetic receiver/channel conditions, the paired correction gain remains positive, although absolute accuracy still degrades under severe CFO and fading. The framework therefore offers a practical basis for cognitive receivers that must exploit complementary models without treating every auxiliary disagreement as a reason to override a reliable primary decision.

\bibliographystyle{IEEEtran}
\bibliography{references_tccn}

\end{document}